\documentclass{article}

    \PassOptionsToPackage{numbers, compress}{natbib}

    \usepackage[preprint]{neurips_2020}

\usepackage[utf8]{inputenc} 
\usepackage[T1]{fontenc}    
\usepackage[pagebackref=true,breaklinks=true,colorlinks,bookmarks=false]{hyperref}       
\usepackage{url}            
\usepackage{booktabs}       
\usepackage{amsfonts}       
\usepackage{nicefrac}       
\usepackage{microtype}      
\usepackage{enumerate}
\usepackage{xcolor}
\usepackage{bm}
\usepackage{graphicx}
\usepackage{booktabs} 
\usepackage{enumitem}
\usepackage{subcaption}
\usepackage{algorithm}
\usepackage[noend]{algpseudocode}
\usepackage{verbatim}
\usepackage{adjustbox}
\usepackage{cleveref}
\usepackage{placeins}
\usepackage{tabularx}
\usepackage{enumitem}
\usepackage{float}
\usepackage{array}
\usepackage{caption}
\usepackage{authblk}

\newcommand*{\ShowNotes}{}

\definecolor{darkred}{rgb}{0.7,0.1,0.1}
\definecolor{darkgreen}{rgb}{0.1,0.7,0.1}
\definecolor{cyan}{rgb}{0.7,0.0,0.7}
\definecolor{dblue}{rgb}{0.2,0.2,0.8}
\definecolor{maroon}{rgb}{0.76,.13,.28}
\definecolor{burntorange}{rgb}{0.81,.33,0}

\ifdefined\ShowNotes
  \newcommand{\colornote}[3]{{\color{#1}\bf{#2: #3}\normalfont}}
\else
  \newcommand{\colornote}[3]{}
\fi

\title{Self-Supervised Dynamic Networks for Covariate Shift Robustness}

\author{Tomer Cohen}
\author{Noy Shulman}
\author{Hai Morgenstern}
\author{Roey Mechrez}
\author{Erez Farhan}
\affil{\texttt{\{tomer.cohen, noy.shulman, hai.morgenstern, roey.mechrez, erez.farhan\}@beyondminds.ai}}
\affil{BeyondMinds}

\begin{document}

\maketitle

\begin{abstract}
As supervised learning still dominates most AI applications, test-time performance is often unexpected. Specifically, a shift of the input covariates, caused by typical nuisances like background-noise, illumination variations or transcription errors, can lead to a significant decrease in prediction accuracy. Recently, it was shown that incorporating self-supervision can significantly improve covariate shift robustness. In this work, we propose Self-Supervised Dynamic Networks (SSDN): an input-dependent mechanism, inspired by dynamic networks, that allows a self-supervised network to predict the weights of the main network, and thus directly handle covariate shifts at test-time. We present the conceptual and empirical advantages of the proposed method on the problem of image classification under different covariate shifts, and show that it significantly outperforms comparable methods.
\end{abstract}

\section{Introduction}
\label{gen_inst}
One of the key limitations of supervised learning is generalization under domain shifts~\cite{quionero2009dataset,moreno2012unifying}, which often leads to significant performance drops when the test distribution is even slightly shifted compared to the train distribution. These cases are highly prevalent in real-life scenarios, especially for smaller training datasets. Thus, a considerable amount of attention is given to the general problem of knowledge transfer between input domains in different scenarios and techniques~\cite{tzeng2017adversarial, hoffman2018cycada, long2017deep, deng2018image, courty2016optimal, wang2018deep}. The vast majority of current works assumes the availability of numerous labeled or unlabeled examples from the target (test) domain ~\cite{deep_domain_conf, deep_coral, ganin2015, ganin2017}, while others may assume access to fewer examples with a possible performance compromise~\cite{motiian2017few, hoffman2013one}. In contrast, when there is access to only a single instance during test phase, subject to an unknown covariate shift, the concept of \textit{target domain} is no longer defined, rendering transfer techniques practically irrelevant. Accordingly, most efforts to solve this problem are focused on detecting data to which the model is not robust, and declaring it out-of-distribution (OOD) ~\cite{OOD_che2019deep,uncerainty_snoek2019can,hendrycks2019benchmarking}. These techniques could avoid prediction errors in cases of a detectable domain shift, but they do not address prediction improvement in these cases. In this work, we examine a specific type of domain shift, known as covariate shift, where only the distribution of the model \emph{inputs} is shifted~\cite{covariate_book,covariate_shift_paper}. We explore methods that go beyond detection, and attempt to improve model prediction on a "covariate-shifted" test sample rather than discarding it.

\begin{figure}[t]
\centering
  \includegraphics[width=1\linewidth]{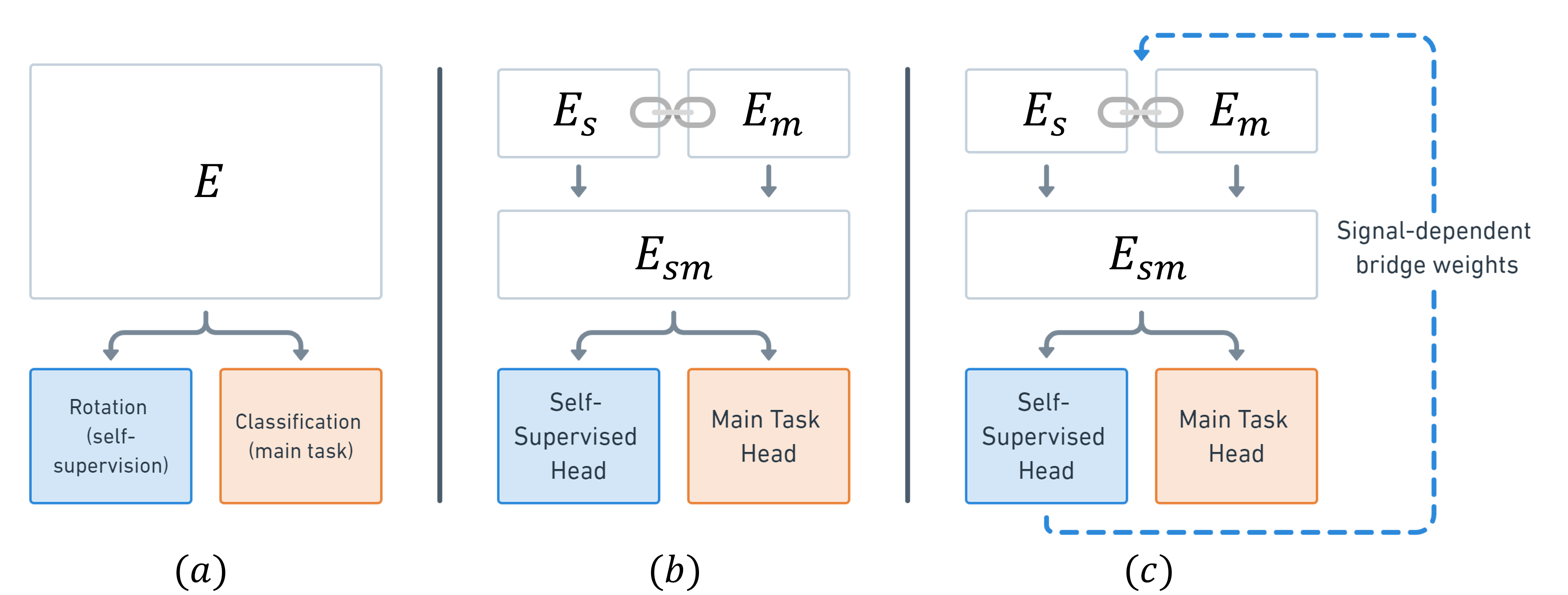}
  \caption{Overview of our approach. (a) Simple shared weights architecture~\cite{hendrycks2019using, sun2019test}. (b) Our method using a data-dependent bridge. $E_s$ is the self-supervised task encoder, and $E_m$ is the main task encoder. The filters of $E_m$ are linear combinations of the filters in $E_s$. (c) Our method using an adaptive, signal-dependent bridge, where the weights for the linear combination are predicted by the self-supervised head, conditioned on the input sample.}
  \label{fig:hypernetworks}
\end{figure}

Specifically for image classification, it was shown that state-of-the-art models, trained with different robustness enhancements, may suffer severe performance degradation in the presence of a simple covariate shift~\cite{hendrycks2019benchmarking}. Recently, Hendrycks et al.~\cite{hendrycks2019using} showed that adding self-supervised learning tasks during training can improve the robustness of visual encoders to these shifts. Sun et al.~\cite{sun2019test} extended this concept to allow for a self-supervised test  time modification of encoder parameters, thus taking a more adjustable approach for achieving robustness. We continue these two lines of work, by addressing both the question of leveraging self-supervised learning to achieve robustness, and the desired types of model adjustability in test time.

A single unlabeled test instance $x$ may contain valuable information about a possible covariate shift. Based on this understanding, allowing the model parameters $\theta$ to depend on $x$ may help adjust the model for such shifts~\cite{sun2019test, zhang2019domain}. Test-time training (TTT), implements this input dependency using a self-supervised mechanism~\cite{sun2019test}. We further improve on this mechanism while suggesting an additional source for input dependency that does not require training at test  time, and significantly improves performance with respect to TTT.

One of questions raised in this paper relates to the components of a model that need to be adjusted to handle covariate shift. Covariate shifts do not typically affect semantic information, and are mostly related to low-level phenomena, which might suggest that robustness should be handled in lower-level processing phases. Specifically for deep neural networks, earlier layers are considered to be more sensitive to low-level information, whereas later layers capture high-level semantic information~\cite{yosinski2014transferable,goodfellow2016deep}. We therefore explore the relationship between the depth of a layer and its robustness to covariate shift, allowing the introduction of more careful model adjustment techniques.

We conduct our experiments in the context of image classification on diverse scenarios and show that our proposed method maintains its performance on the original domain while making substantial improvements under covariate shift with respect to previous work. Our key contributions are:
\begin{enumerate}[]
    \item The first framework to explicitly translate self-supervised neural representations to a main task representation, using a direct differentiable mapping approach.
    \item A novel approach for using input-dependent networks to alleviate test-time performance degradation due to covariate shift, and a demonstration of its superiority over competing approaches.
    \item Pushing the envelope of the relatively unexplored idea to achieve robustness through test-time adjustments.
    \item An analysis tool for determining which CNN components require modifications in order to better cope with covariate shifts, and using it to provide a lucid justification for our layer-specific approach.
\end{enumerate}

\section{Related work}
\subsection{Self-supervised learning}
The field of self-supervised representation learning has made great strides recently, with massive success in language modeling~\cite{BERT,GPT2}, speech~\cite{CPC,PASE} and very recently in computer vision~\cite{CPC2,SIMCLR,MoCo}. 
In this work, we focus on a simple yet effective self-supervised technique for learning visual representations~\cite{ss_rotation}, and note that this approach can be extended to other fields and arbitrary (appropriate) self-supervision techniques.
In the context of covariate shift, it was recently shown that training an encoder in a multi-task fashion alongside a self-supervised task can result in more robust image classification performance~\cite{hendrycks2019using, sun2019test} (see Fig.~\ref{fig:hypernetworks}-a). We show that a fundamental limitation of ~\cite{hendrycks2019using, sun2019test} comes from strictly sharing the encoder between the self-supervised and main task. In contrast, we propose a direct approach for optimizing knowledge sharing between the self-supervised encoder and a \emph{separate} main task encoder.

\subsection{Input-dependent neural networks}
Our proposed input-dependent method is inspired by the idea of hyper or dynamic networks~\cite{bertinetto2016learning, kristiadiuncertainty, nachmani2019hyper}, where the parameters of one neural network can be determined by the output of another, applied to the same input. We extend this concept to have a self-supervised encoder dynamically control the parameters of a main task encoder.

\subsection{Data augmentation}
It has been recently shown that massive data augmentation can improve model robustness to covariate shift in different scenarios and fields ~\cite{augmix,autoaugment,specaugment,nlp_augment_consistency}.
While this approach is highly applicable in many cases, it requires significant inductive bias which may not be available for every task. In contrast, this work addresses modifications in the learning model, and is therefore agnostic to the data or the task. Additionally, it does not carry the computational overhead introduced by training with data augmentations.
Thus, we leave data augmentation techniques outside the scope of this work.
\section{Problem formulation}
\label{problem}
For a supervised learning model $y=F_\theta(x)$, with parameters $\theta$, a covariate shift occurs where the input distribution changes between training and test  time (i.e. $P_{test}(X) \neq P_{train}(X)$), while the conditional distribution ($P(Y|X)$) is preserved. We also assume that $P_{test}(X)$ is unknown. Our goal is to have the learning model be robust to this shift, such that for a given pair $(x,y)| x \sim P_{test}$, we will obtain:
\begin{equation}
    F_\theta(x) \approx y
\end{equation}
Specifically in this work, we are interested in a family of solutions that have the parameters $\theta$ depend on the input. That is $F_{\theta}(x)=F_{\theta(x)}(x)$.
\section{Background}
\label{background}

\subsection{Joint training}
Joint training (JT) employs a reversed Y-shaped architecture to train a two-head neural network, with a \emph{shared feature extractor}~\cite{hendrycks2019using}. One head is for a self-supervised task (e.g. rotation angle prediction), and the second head is for the main downstream task (e.g. classification). See Fig.~\ref{fig:hypernetworks} for a schematic diagram of this architecture. We denote \boldsymbol{$\theta_e$} the parameters of the shared feature extractor $E$, \boldsymbol{$\theta_s$} the parameters of the self-supervised head $S$, and \boldsymbol{$\theta_m$} the parameters of the main task head $M$. At training time, the self-supervised loss $l_s$ and the main-task loss $l_m$, are jointly optimized using data from a distribution $P_{train}$. At test time, the model only employs the main task head, while the self-supervised head is discarded.
\subsection{Test-time training}
\label{baseline}
 TTT~\cite{sun2019test} is a modification of JT that enables the shared encoder to also learn at test time, without the main task labels. Specifically, the self-supervised loss $l_s$ is minimized using the test example $x_t$ and yields the approximated minimizer, $\boldsymbol{\theta_e^*}$, of:
 \begin{equation}
\label{eq:test-time}
\displaystyle \min_{\boldsymbol{\theta_e}, \boldsymbol{\theta_s}} l_s(x_t,\boldsymbol{\theta_e},\boldsymbol{\theta_s})
\end{equation} Then, the main-task prediction is done using the updated parameters:
\begin{equation}
\label{eq:test-time-predict}
\displaystyle 
\boldsymbol{\theta}(x_t) = (\boldsymbol{\theta_e^*},\boldsymbol{\theta_m})
\end{equation}
Noting that the model parameters $\theta$ now depend on the test input $x_t$.

Both the JT and its TTT extension assume that the main-task prediction somehow improves by optimizing the shared-encoder \boldsymbol{$\theta_e$} to minimize the self-supervised loss $l_s$. Though this assumption might be justified by regularization considerations, it seems that the usage of self-supervision to benefit downstream tasks can be done more systematically both in training and in test time.
Another point regards the dependency of the model parameters in the test input $x_t$. In TTT, this is done purely through back-propagation from $l_s$. This input dependency procedure is empirically backed in~\cite{sun2019test}, demonstrating correlated gradients of $l_m$ and $l_s$, with respect to \boldsymbol{$\theta_e$}. However, it is not directly being optimized during training. In contrast,
we observe that introducing input dependency through the model architecture may allow for a direct optimization of input conditioning.
In the next section, we propose two extensions to JT that directly address these issues, and describe how they can also be combined with the TTT principle.
\section{Test-time adjustments using dynamic neural networks}
\label{method}  
Our first observation is that the different layers of deep network may have different sensitivity to covariate shifts. For example, since earlier layers capture low-level information, it may be beneficial to focus our effort on them. Consequently, we propose a \textit{split architecture} as illustrated in Fig.~\ref{fig:hypernetworks}-b, where we split the first layers of the shared encoder $E$ of JT to be separate for the main task ($E_m$) and the self-supervised task ($E_s$). Thus, allowing us to customize their behaviour, while the rest of the encoder ($E_{sm}$) remains shared as in JT.
Next, we address the issue of a more systematic usage of self-supervised representations with the split architecture (Sec. \ref{data_dependent_bridge}), and continue with proposing an architectural adjustment that allows it to be explicitly optimized on the test sample (Sec. \ref{signal_dependent_bridge}).
\subsection{Data-dependent bridge}
\label{data_dependent_bridge}
Using the split architecture, we wish to optimize the task module $E_m$, using the self-supervised module $E_s$, by mapping the weights of $E_s$ to determine those of $E_m$. For this, every convolutional layer $C^j_s\in E_s$ is used as a filter bank, such that every filter in a corresponding layer $C^i_m\in E_m$, with parameters $w_i^m$, is a weighted linear combination of the filter parameters $w_j^s$ of $C^j_s$, as follows:
\begin{equation}
w_i^m = \sum_{j=1}^{J} \alpha^d_{ij} w_j^s
\end{equation}
where $J$ is the number of convolutional filters in $C^i_m$, and $\{\alpha^d_{ij}\}^{I, J}_{i, j = 1}$ are learned, \textit{data-dependent} parameters, optimized in the training stage.
This unique connection between $E_m$ and $E_s$ can be seen as a "bridge" between the filters of the self-supervised task and the main task. Thus, this architecture is able to leverage covariate shift information acquired by $E_s$, to benefit the main task. Since this bridge is learned end-to-end against all the training data, we term it \textit{data-dependent bridge}.

\subsection{Signal-dependent bridge}
\label{signal_dependent_bridge}
Following the mapping motivation of the data-dependent bridge, we propose a similar \textit{signal-dependent bridge}. In this setting, the weighted combination for $w_i^m$ is conditioned on the input signal. Specifically, the self-supervision head now predicts both the self-supervised labels and the bridge parameters $ \{ \alpha^s_{ij}(x_t) \}_{i, j=1}^{I, J}$, which are now \textit{signal-dependent}. This approach defines a conditioning between the input signal and the main task network. To combine this setting with the data-dependent bridge, we simply sum the filters predicted by both bridges. That is:
\begin{equation}
w_i^m = \sum_{j=1}^{J} \left(\alpha^s_{ij}(x_t) + \alpha^d_{ij}\right) w_j^s
\end{equation}
We note that the presented methods are complementary with respect to their usage in self-supervised learning:
the first (data-dependent) relies on the quality of the self-supervised representation, and statically maps it to the main representation, while the second (signal-dependent) relies on tight tuning of the self-supervised network to the specific input, and thus allows for a more dynamic adjustment.

We note that since our method is an extension to JT and is fully differentiable, it can also benefit from the TTT principle, which introduces a complementary input-dependent signal. We thus examine this combination in our experimental study (Sec.~\ref{single-instance-experiment}).

\section{Analysis of covariate shift sensitivity}
\label{theoretical}

\begin{figure}[H]
\begin{subfigure}{.5\textwidth}
  \includegraphics[width=1.0\linewidth]{cka.png}
   \caption{Block-wise fine tuning}
   \label{fig:CKA experiment}
\end{subfigure}
\begin{subfigure}{0.5\textwidth}
  \centering
  \includegraphics[width=1.0\linewidth]{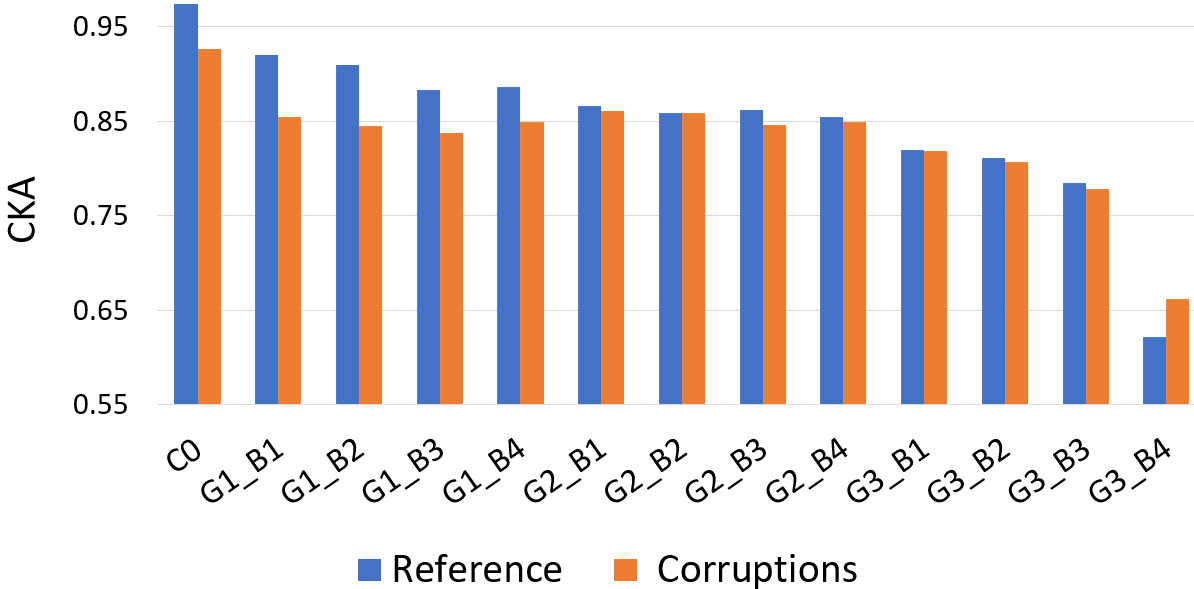}
  \caption{Similarity scores per block}
  \label{fig:representation_similarity_freeze}
\end{subfigure}

\caption{Covariate shift analysis. (\ref{fig:CKA experiment}) CKA index compares a specific layer of a reference encoder with the same encoder find-tuned on shifted data. Green: fine-tuned block. Gray: frozen. (\ref{fig:representation_similarity_freeze}) CKA index per layer. Early layers are more sensitive to covariate shift.
}
\end{figure}

Throughout all the experiments in this section and in Sec.~\ref{single-instance-experiment}, we utilize the same ResNet-26 architecture as used by \cite{sun2019test} for the example task of image classification with the exact same hyperparameters. As illustrated in Fig.~\ref{fig:CKA experiment}, a ResNet-26 encoder is composed of an initial convolutional layer (C0), followed by four groups (G1-G4), containing four residual blocks (B1-B4) each. In this section, we present an experiment for analyzing the sensitivity of this specific architecture to covariate shifts. 

\subsection{Setup} 
We conduct the experiment on the CIFAR-10-C dataset~\cite{hendrycks2019benchmarking}, treating the original CIFAR-10~\cite{krizhevsky2009learning} as reference, and the corrupted versions as target datasets simulating different covariate shifts. All networks are trained for image classification using a standard supervised-learning procedure.

\subsection{Methods} 
To analyze the sensitivity of each network component to covariate shift, we developed a method for isolating the effect of covariate shift on the functionality of each block. As illustrated in Fig.~\ref{fig:CKA experiment}, the following procedure is conducted:
First, an encoder is trained using a reference dataset (reference encoder). Then, given a covariate-shifted dataset, the reference encoder is fine-tuned, allowing a specific block to optimize while all other blocks are kept frozen. Finally, a representation similarity index is computed between the reference and tuned block. Repeating this experiment across different blocks and different covariate shifts, enables to analyze the covariate shift sensitivity of each layer.

For control, the similarity index is also computed between multiple reference encoders trained from different random initializations. The similarity score of a fine-tuned encoder and a reference encoder is expected to be lower than the score of two reference encoders.

To measure representation similarity, the Centered Kernel Alignment (CKA) index~\cite{CKA} is employed. A low CKA index implies low similarity between blocks, indicating a higher sensitivity of the specific block to the covariate shift. To get a reliable CKA measure, multiple experiments are ran for each CIFAR-10-C corruption type.

\subsection{Results}
The representation similarity results for each block is presented in Fig.~\ref{fig:representation_similarity_freeze}. We were particularly interested in the blocks where the baseline similarity was kept high and above the similarity under corruptions. Clearly, and consistent with~\cite{yosinski2014transferable}, the largest gap was observed in earlier layers, indicating their sensitivity to covariate shift. This behavior was generally consistent across the different corruptions. This suggests that earlier layers are indeed more sensitive to covariate shift, and that their adjustment is more crucial in this context. Note that although we observed significant sensitivity of the first convolutional layer (C0), we found that adjusting this layer with our proposed method resulted in training instability. We thus avoided adjusting this layer in the following experiments, and only adjusted the blocks in G1. The respective ablation study is given in Sec.~\ref{ablation}.

\section{Empirical Analysis}
\label{single-instance-experiment}

\subsection{Setup} 
We validate SSDN on a series of image classification experiments, conducted on different datasets. In all experiments, all the methods were trained against a source dataset, and tested against a dataset containing different covariate shifts. We assumed that the test instances went through independent covariate shifts with respect to the training set distribution. This setup is termed "Single".
In particular, the following configurations were examined using the "Single" setup:
\begin{enumerate}[leftmargin=1cm]
    \item CIFAR-10$\rightarrow$CIFAR-10-C: As in ~\cite{hendrycks2019benchmarking,sun2019test}, CIFAR-10~\cite{krizhevsky2009learning} was used as the source domain, while the corrupted versions from CIFAR-10-C simulated the different covariate shifts~\cite{hendrycks2019benchmarking}.
    \item CIFAR-10$\rightarrow$CIFAR-10.1: Similarly, CIFAR-10 was used as the source domain, while CIFAR-10.1 simulated the covariate shift~\cite{recht2018cifar}. This case is particularly interesting, since the covariate shift from CIFAR-10 to CIFAR10.1 is known to be very challenging to capture~\cite{recht2018cifar}.
    \item SVHN$\rightarrow$MNIST: In this experiment, SVHN~\cite{svhn} was used as the source dataset, while covariate shifted samples were taken from the MNIST~\cite{mnist} dataset. 
\end{enumerate}
 
 An additional experiment is included, where unlabeled test instances arrive online and are assumed to come from the same distribution. This setup is termed "Online". As there are different approaches to tackle this unsupervised online problem, we note that this experiment is not complete with respect to the possible state-of-the-art, but is provided solely for compliance with ~\cite{sun2019test}. Specifically, as the online queue increases, comparison to domain-adaptation methods is due.

\subsection{Methods}
The following baseline methods are compared, all based on the same ResNet-26 architecture:
\begin{itemize}[leftmargin=*]
  \setlength\itemsep{0em}
  \item[] \textbf{Standard} - A supervised model trained with classification labels.
  \item[] \textbf{Joint training} - Combining supervised and self-supervised learning using rotation estimation on a shared encoder \cite{hendrycks2019using}.
  \item[] \textbf{Original TTT} - Performing test-time training on top of "Joint training" \cite{sun2019test}.
  \item[] \textbf{SSDN one-pass} - Splitting G1 into $E_s$ and $E_m$ and combining the data-dependent and the signal-dependent bridge as described in Sec.~\ref{method}.
  \item[] \textbf{SSDN + TTT} - Performing test-time training on top of "SSDN one-pass".
\end{itemize}
In all TTT experiments, we ran $K=16$ backprop iterations to minimize $l_s$ before final classification. All the other hyperparameters were identical to \cite{sun2019test}.

\label{CIFAR-10-C_exp}

\setlength\belowcaptionskip{-2ex}

\begin{figure}[h]
\centering
  \makebox[\textwidth][c]{\includegraphics[width=1.01\linewidth]{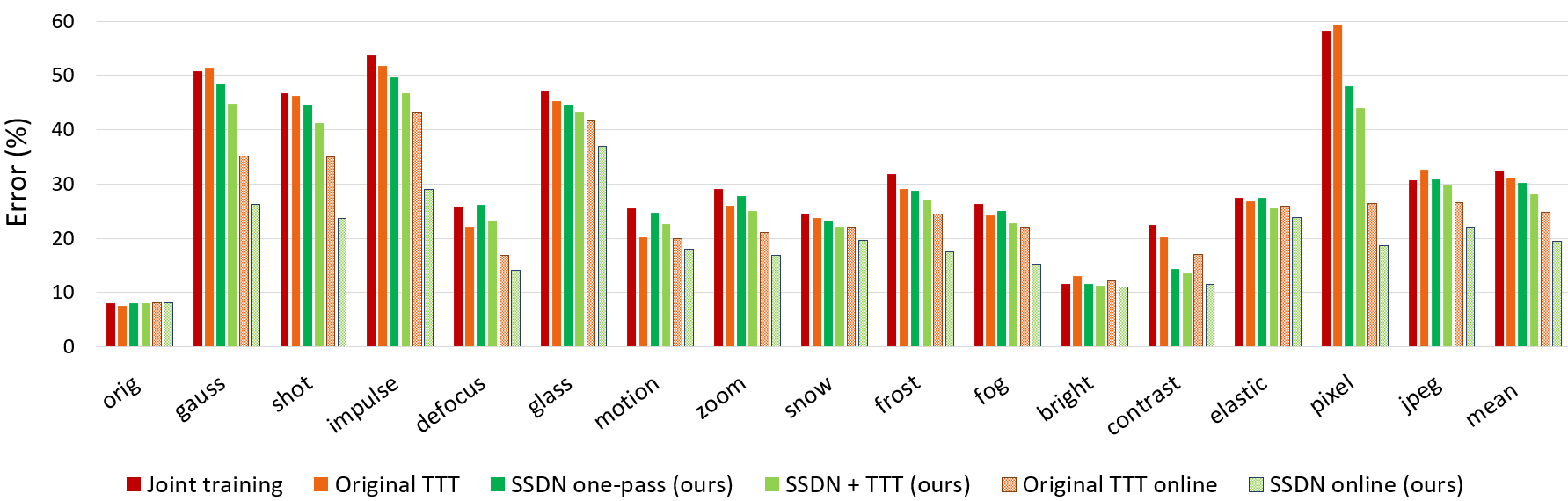}}
  \caption{Test errors on {\small CIFAR-10-C} dataset (best viewed in color). {\small SSDN} outperforms the baselines even at a one feed-forward scenario. Combining our method with {\small TTT} further improves performance.}
  \label{fig:cifar10c}
\end{figure}

\subsection{Results}
\textbf{CIFAR-10$\rightarrow$CIFAR-10-C. }Fig.~\ref{fig:cifar10c} depicts the experiments comparing SSDN to the baseline methods on the CIFAR-10-C dataset. Evidently, this pure feed-forward variant of SSDN performed better than the baseline TTT, making it a very attractive variant for real-time applications. Combining test-time training with SSDN further improves accuracy and achieves 9.7\% relative improvement over the closest baseline. The results for the "Online" versions of the baseline TTT and SSDN expectedly outperformed their "Single" counterparts, indicating that this setup is significantly less challenging. In this setup, SSDN outperformed~\cite{sun2019test} by a large margin of 21.4\% relative improvement.

\textbf{CIFAR-10$\rightarrow$CIFAR-10.1. }A similar trend was observed in the CIFAR-10.1 experiment (Tab.~\ref{cifar101}),  where both our one-pass and TTT versions significantly outperformed all baselines. Our TTT variant improves by 10.6\% over \cite{sun2019test} and 15\% over \cite{hendrycks2019using}. Consistent with ~\cite{recht2018cifar}, these results might explain the poor generalization of standard CIFAR-10 models to CIFAR-10.1. Namely, the existence of covariate shifts. It seems that our input-dependent model was able to generalize from the internal covariate shift exists in the CIFAR-10 dataset to better handle some of the variations presented by the CIFAR-10.1 dataset.

\textbf{SVHN$\rightarrow$MNIST. }Tab.~\ref{mnist_svhn} shows results for SVHN $\rightarrow$ MNIST experiment. Our "Single" variant outperforms all baselines by at least 14\% relative improvement. Interestingly, both SSDN and JT~\cite{hendrycks2019using} do not gain further improvement after running TTT~\cite{sun2019test} on this scenario. We hypothesise this may be due to the large domain gap in this experiment. While an initial bridging improves performance by a considerable margin, TTT might overfit the sample, resulting in bridge parameters which are too far from the original distribution, and thus degrades performance.

When running the experiment the other way around (MNIST $\rightarrow$ SVHN), all methods performed very poorly. SVHN has much larger sample variance than MNIST, thus generalization in this direction is far from trivial and out of reach for the compared methods.


\begin{table}[H]
\parbox{.45\linewidth}{
\centering
\caption{{\small CIFAR-10.1} test errors.}
\label{cifar101}
\begin{tabular}{@{}lc@{}}
 \toprule
Method & Error (\%)  \\ 
\midrule 
Standard & 17.4  \\ 
Joint training \cite{hendrycks2019using} & 16.7  \\ 
Original \small{TTT} \cite{sun2019test} & 15.9  \\ 
SSDN one-pass & 14.6  \\ 
SSDN + \small{TTT} & \textbf{14.2}  \\ 
\bottomrule
\end{tabular}
}
\hfill
\parbox{.45\linewidth}{
\centering
\caption{{\small SVHN$\rightarrow$MNIST} test errors.}
\label{mnist_svhn}
\begin{tabular}{@{}lc@{}}

 \toprule
Method & Error (\%)  \\ 
\midrule 
Standard & 41.9 \\ 
Joint training \cite{hendrycks2019using} & 29.6  \\ 
Original \small{TTT} \cite{sun2019test} & 29.7  \\ 
SSDN one-pass & \textbf{25.4} \\ 
SSDN + \small{TTT} & 25.7  \\ 
\bottomrule
\end{tabular}
}
\end{table}

\subsection{Ablation study}
\label{ablation}
To better understand the effect of bridging between $E_s$ and $E_m$ at different layers, we evaluated different versions of SSDN, where each time we used different parts of the encoder as the bridge (see Tab.~\ref{block_ablation}). While bridging all layers may sound like the optimal scheme from an optimization perspective, there seems to be an important inductive bias related to the sensitivity of each block to covariate shifts (see Sec.~\ref{fig:representation_similarity_freeze}), and specifically in focusing on the lower level part of the encoder (G1). It also seems that optimizing the bridge is most stable when choosing single groups rather than mixing them.

\subsection{Analysis of the bridge parameters}
\label{Alpha-Clustering-Experiment}

\begin{figure}[h]
\centering
  \includegraphics[width=1\linewidth]{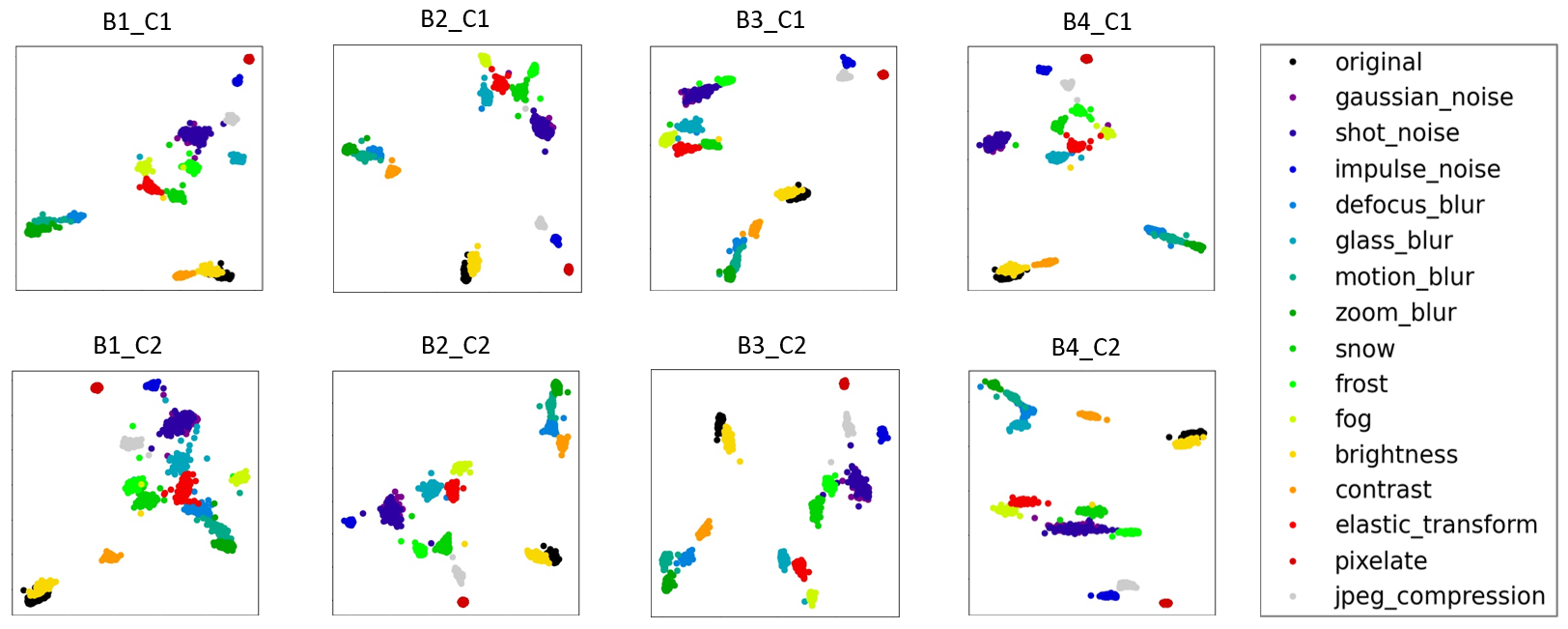}
  \caption{{$\alpha^s$ clustering}. t-{\small SNE} plots of the bridge parameters ($\alpha^s$), predicted by the self-supervised head. $B[i]\_C[j]$ refers to the $j^{th}$ conv layer in the $i^{th}$ residual block of G1. "original" refers to the CIFAR-10 test set (best viewed in color).}
  \label{fig:alphas}
\end{figure}

To directly analyze the effect of the proposed signal-dependent bridge (Sec.~\ref{signal_dependent_bridge}), we inspected the $\alpha^s$ parameters under different covariate shifts from the CIFAR-10-C experiment under the "one-pass" variant. Results are visualized in Fig.~\ref{fig:alphas}, where distinct clusters are prominent for almost every covariate shift. Thus, it seems that the signal-dependent bridge was able to detect the covariate shift in a completely unsupervised manner, and therefore allowed the encoder to adapt accordingly.

Interestingly, the cluster location on the plot is a good proxy for the classification accuracy under the respective covariate shift. For example, "brightness" is the closest to "original", and both have the lowest test error. In contrast, "gaussian\_noise" is fairly far from "original", and has a very high test error. Moreover, when looking at the t-SNE plots of the different layers, the clusters seem to behave isometrically, further suggesting that the model gains valuable and consistent covariate shift information from each sample, in an unsupervised fashion.

\vspace*{-\baselineskip}

\begin{table}[H]

\centering
\caption{Bridging different encoders parts.}
\label{block_ablation}

\begin{tabular}{@{}ccccc@{}}
\toprule
{\small C0} & {\small G1} & {\small G2} & one-pass & one-pass+{\small TTT}  \\ 
\midrule 
0 & 0 & 1 & 30.71 & 28.69  \\ 
0 & 1 & 0 & \textbf{30.21} & \textbf{28.18} \\ 
0 & 1 & 1 & 32.16 & 30.08 \\ 
1 & 0 & 0 & 30.46 & 28.57 \\ 
1 & 0 & 1 & 33.95 & 31.64 \\ 
1 & 1 & 0 & 31.05 & 29.07 \\ 
1 & 1 & 1 & 32.53 & 30.26 \\ 
\bottomrule
\end{tabular}
\end{table}
\vspace*{-\baselineskip}

\subsection{Discussion}
The results presented in this study demonstrate the contribution of self-supervised learning for handling covariate shifts, with all compared variants surpassing the equivalent fully supervised model. We observed how a direct optimization of this contribution significantly improved image classification performance on a variety of scenarios, perhaps calling for further experimentation with the presented ideas on different problems and other self-supervision techniques.
 In light of the analysis in \ref{Alpha-Clustering-Experiment}, the prominent empirical performance of the proposed input-dependent mechanism might be attributed to the hierarchical nature of this architecture. It seems that the model was able to first gather information about the covariate shift, and then use it for a more appropriate analysis of the input. Such input conditional behaviour can prove crucial when dealing with significant intra-dataset variations, or equivalently for dealing with catastrophic forgetting in continual learning or online setups.

\section{Conclusions}
In this paper, the problem of model robustness to covariate shift was addressed. We proposed a sensitivity analysis to better pinpoint the sources of non-robustness of CNNs. Motivated by the analysis, we introduced methods for optimizing the relationship between self-supervised representations and main task representations, and for direct input-dependent adjustment to covariate shift. We demonstrated the attractiveness of our methods in several image classification scenarios versus recently proposed alternatives. With some architectural modifications, we believe these proposed ideas can be applied to other machine learning domains to improve robustness with test-time adjustments. Specifically, we believe that the proposed input conditional mapping between representations can be extended to tackle other challenges beyond covariate shift robustness, such as online domain adaptation and continual learning, and can contribute to the growing impact of self-supervised learning techniques.


\bibliographystyle{unsrt}
\bibliography{refs}

\begin{thebibliography}{10}

\bibitem{quionero2009dataset}
Joaquin Quionero-Candela, Masashi Sugiyama, Anton Schwaighofer, and Neil~D
  Lawrence.
\newblock {\em Dataset shift in machine learning}.
\newblock The MIT Press, 2009.

\bibitem{moreno2012unifying}
Jose~G Moreno-Torres, Troy Raeder, Roc{\'\i}O Alaiz-Rodr{\'\i}Guez, Nitesh~V
  Chawla, and Francisco Herrera.
\newblock A unifying view on dataset shift in classification.
\newblock {\em Pattern recognition}, 45(1):521--530, 2012.

\bibitem{tzeng2017adversarial}
Eric Tzeng, Judy Hoffman, Kate Saenko, and Trevor Darrell.
\newblock Adversarial discriminative domain adaptation.
\newblock In {\em Proceedings of the IEEE Conference on Computer Vision and
  Pattern Recognition}, 2017.

\bibitem{hoffman2018cycada}
Judy Hoffman, Eric Tzeng, Taesung Park, Jun-Yan Zhu, Phillip Isola, Kate
  Saenko, Alexei Efros, and Trevor Darrell.
\newblock {CyCADA}: Cycle-consistent adversarial domain adaptation.
\newblock In {\em International Conference on Machine Learning}, 2018.

\bibitem{long2017deep}
Mingsheng Long, Han Zhu, Jianmin Wang, and Michael~I Jordan.
\newblock Deep transfer learning with joint adaptation networks.
\newblock In {\em Proceedings of International Conference on Machine Learning},
  2017.

\bibitem{deng2018image}
Weijian Deng, Liang Zheng, Qixiang Ye, Guoliang Kang, Yi~Yang, and Jianbin
  Jiao.
\newblock Image-image domain adaptation with preserved self-similarity and
  domain-dissimilarity for person re-identification.
\newblock In {\em Proceedings of the IEEE conference on computer vision and
  pattern recognition}, 2018.

\bibitem{courty2016optimal}
Nicolas Courty, R{\'e}mi Flamary, Devis Tuia, and Alain Rakotomamonjy.
\newblock Optimal transport for domain adaptation.
\newblock {\em IEEE transactions on pattern analysis and machine intelligence},
  2016.

\bibitem{wang2018deep}
Mei Wang and Weihong Deng.
\newblock Deep visual domain adaptation: A survey.
\newblock {\em Neurocomputing}, 2018.

\bibitem{deep_domain_conf}
Ning Zhang Kate~Saenko Eric~Tzeng, Judy~Hoffman and Trevor Darrell.
\newblock Contrastive adaptation network for unsupervised domain adaptation.
\newblock In {\em arXiv preprint arXiv:1412.3474}, 2014.

\bibitem{deep_coral}
Baochen Sun and Kate Saenko.
\newblock Deep coral: Correlation alignment for deep domain adaptation.
\newblock In {\em 2016 European Conference on Computer Vision}, 2016.

\bibitem{ganin2015}
Yaroslav Ganin and Victor~S. Lempitsky.
\newblock Unsupervised domain adaptation by backpropagation.
\newblock In {\em the 32nd International Conference on Machine Learning
  (ICML)}, 2015.

\bibitem{ganin2017}
Hana Ajakan Pascal Germain Hugo Larochelle Mario~Marchand Yaroslav~Ganin,
  Evgeniya~Ustinova and Victor Lempitsky.
\newblock Domain-adversarial training of neural networks.
\newblock In {\em Journal of Machine Learning Research}, volume~17, page
  2096–2030, 2017.

\bibitem{motiian2017few}
Saeid Motiian, Quinn Jones, Seyed Iranmanesh, and Gianfranco Doretto.
\newblock Few-shot adversarial domain adaptation.
\newblock In {\em Advances in Neural Information Processing Systems}, 2017.

\bibitem{hoffman2013one}
Judy Hoffman, Eric Tzeng, Jeff Donahue, Yangqing Jia, Kate Saenko, and Trevor
  Darrell.
\newblock One-shot adaptation of supervised deep convolutional models.
\newblock {\em arXiv preprint arXiv:1312.6204}, 2013.

\bibitem{OOD_che2019deep}
Tong Che, Xiaofeng Liu, Site Li, Yubin Ge, Ruixiang Zhang, Caiming Xiong, and
  Yoshua Bengio.
\newblock Deep verifier networks: Verification of deep discriminative models
  with deep generative models.
\newblock {\em arXiv preprint arXiv:1911.07421}, 2019.

\bibitem{uncerainty_snoek2019can}
Jasper Snoek, Yaniv Ovadia, Emily Fertig, Balaji Lakshminarayanan, Sebastian
  Nowozin, D~Sculley, Joshua Dillon, Jie Ren, and Zachary Nado.
\newblock Can you trust your model's uncertainty? evaluating predictive
  uncertainty under dataset shift.
\newblock In {\em Advances in Neural Information Processing Systems}, pages
  13969--13980, 2019.

\bibitem{hendrycks2019benchmarking}
Dan Hendrycks and Thomas Dietterich.
\newblock Benchmarking neural network robustness to common corruptions and
  perturbations.
\newblock {\em arXiv preprint arXiv:1903.12261}, 2019.

\bibitem{covariate_book}
Masashi Sugiyama and Motoaki Kawanabe.
\newblock {\em Machine learning in non-stationary environments: Introduction to
  covariate shift adaptation}.
\newblock 2012.

\bibitem{covariate_shift_paper}
Hidetoshi Shimodaira.
\newblock Improving predictive inference under covariate shift by weighting the
  log-likelihood function.
\newblock {\em Journal of statistical planning and inference}, 90(2):227--244,
  2000.

\bibitem{hendrycks2019using}
Dan Hendrycks, Mantas Mazeika, Saurav Kadavath, and Dawn Song.
\newblock Using self-supervised learning can improve model robustness and
  uncertainty.
\newblock In {\em Advances in Neural Information Processing Systems}, pages
  15637--15648, 2019.

\bibitem{sun2019test}
Yu~Sun, Xiaolong Wang, Zhuang Liu, John Miller, Alexei~A Efros, and Moritz
  Hardt.
\newblock Test-time training for out-of-distribution generalization.
\newblock {\em arXiv preprint arXiv:1909.13231}, 2019.

\bibitem{zhang2019domain}
Tianyuan Zhang, Bichen Wu, Xin Wang, Joseph Gonzalez, and Kurt Keutzer.
\newblock Domain-aware dynamic networks.
\newblock {\em arXiv preprint arXiv:1911.13237}, 2019.

\bibitem{yosinski2014transferable}
Jason Yosinski, Jeff Clune, Yoshua Bengio, and Hod Lipson.
\newblock How transferable are features in deep neural networks?
\newblock In {\em Advances in neural information processing systems}, pages
  3320--3328, 2014.

\bibitem{goodfellow2016deep}
Ian Goodfellow, Yoshua Bengio, and Aaron Courville.
\newblock {\em Deep learning}.
\newblock MIT press, 2016.

\bibitem{BERT}
Jacob Devlin, Ming-Wei Chang, Kenton Lee, and Kristina Toutanova.
\newblock Bert: Pre-training of deep bidirectional transformers for language
  understanding.
\newblock {\em arXiv preprint arXiv:1810.04805}, 2018.

\bibitem{GPT2}
Alec Radford, Jeffrey Wu, Rewon Child, David Luan, Dario Amodei, and Ilya
  Sutskever.
\newblock Language models are unsupervised multitask learners.
\newblock {\em OpenAI Blog}, 1(8):9, 2019.

\bibitem{CPC}
Aaron van~den Oord, Yazhe Li, and Oriol Vinyals.
\newblock Representation learning with contrastive predictive coding.
\newblock {\em arXiv preprint arXiv:1807.03748}, 2018.

\bibitem{PASE}
Santiago Pascual, Mirco Ravanelli, Joan Serr{\`a}, Antonio Bonafonte, and
  Yoshua Bengio.
\newblock Learning problem-agnostic speech representations from multiple
  self-supervised tasks.
\newblock {\em arXiv preprint arXiv:1904.03416}, 2019.

\bibitem{CPC2}
Olivier~J H{\'e}naff, Aravind Srinivas, Jeffrey De~Fauw, Ali Razavi, Carl
  Doersch, SM~Eslami, and Aaron van~den Oord.
\newblock Data-efficient image recognition with contrastive predictive coding.
\newblock {\em arXiv preprint arXiv:1905.09272}, 2019.

\bibitem{SIMCLR}
Ting Chen, Simon Kornblith, Mohammad Norouzi, and Geoffrey Hinton.
\newblock A simple framework for contrastive learning of visual
  representations.
\newblock {\em arXiv preprint arXiv:2002.05709}, 2020.

\bibitem{MoCo}
Kaiming He, Haoqi Fan, Yuxin Wu, Saining Xie, and Ross Girshick.
\newblock Momentum contrast for unsupervised visual representation learning.
\newblock {\em arXiv preprint arXiv:1911.05722}, 2019.

\bibitem{ss_rotation}
Praveer~Singh Spyros~Gidaris and Nikos Komodakis.
\newblock Unsupervised representation learning by predicting image rotations.
\newblock In {\em ICLR}, 2018.

\bibitem{bertinetto2016learning}
Luca Bertinetto, Jo{\~a}o~F Henriques, Jack Valmadre, Philip Torr, and Andrea
  Vedaldi.
\newblock Learning feed-forward one-shot learners.
\newblock In {\em Advances in neural information processing systems}, pages
  523--531, 2016.

\bibitem{kristiadiuncertainty}
Agustinus Kristiadi, Sina D{\"a}ubener, and Asja Fischer.
\newblock Uncertainty quantification with compound density networks.

\bibitem{nachmani2019hyper}
Eliya Nachmani and Lior Wolf.
\newblock Hyper-graph-network decoders for block codes.
\newblock In {\em Advances in Neural Information Processing Systems}, 2019.

\bibitem{augmix}
Dan Hendrycks, Norman Mu, Ekin~D. Cubuk, Barret Zoph, Justin Gilmer, and Balaji
  Lakshminarayanan.
\newblock {AugMix}: A simple data processing method to improve robustness and
  uncertainty.
\newblock {\em Proceedings of the International Conference on Learning
  Representations (ICLR)}, 2020.

\bibitem{autoaugment}
Ekin~D Cubuk, Barret Zoph, Dandelion Mane, Vijay Vasudevan, and Quoc~V Le.
\newblock Autoaugment: Learning augmentation strategies from data.
\newblock In {\em Proceedings of the IEEE conference on computer vision and
  pattern recognition}, pages 113--123, 2019.

\bibitem{specaugment}
Daniel~S Park, William Chan, Yu~Zhang, Chung-Cheng Chiu, Barret Zoph, Ekin~D
  Cubuk, and Quoc~V Le.
\newblock Specaugment: A simple data augmentation method for automatic speech
  recognition.
\newblock {\em arXiv preprint arXiv:1904.08779}, 2019.

\bibitem{nlp_augment_consistency}
Qizhe Xie, Zihang Dai, Eduard Hovy, Minh-Thang Luong, and Quoc~V Le.
\newblock Unsupervised data augmentation for consistency training.
\newblock 2019.

\bibitem{krizhevsky2009learning}
Alex Krizhevsky, Geoffrey Hinton, et~al.
\newblock Learning multiple layers of features from tiny images.
\newblock 2009.

\bibitem{CKA}
Simon Kornblith, Mohammad Norouzi, Honglak Lee, and Geoffrey Hinton.
\newblock Similarity of neural network representations revisited.
\newblock {\em arXiv preprint arXiv:1905.00414}, 2019.

\bibitem{recht2018cifar}
Benjamin Recht, Rebecca Roelofs, Ludwig Schmidt, and Vaishaal Shankar.
\newblock Do {CIFAR}-10 classifiers generalize to {CIFAR}-10?
\newblock {\em arXiv preprint arXiv:1806.00451}, 2018.

\bibitem{svhn}
Pierre Sermanet, Soumith Chintala, and Yann LeCun.
\newblock Convolutional neural networks applied to house numbers digit
  classification.
\newblock In {\em Proceedings of International Conference on Pattern
  Recognition}. IEEE, 2012.

\bibitem{mnist}
Yann LeCun, L{\'e}on Bottou, Yoshua Bengio, and Patrick Haffner.
\newblock Gradient-based learning applied to document recognition.
\newblock {\em Proceedings of the IEEE}, 86(11):2278--2324, 1998.

\end{thebibliography}

\end{document}